\documentclass[runningheads]{llncs}
\usepackage{graphicx}
\usepackage[colorlinks]{hyperref}
\usepackage{listings}
\usepackage{courier}
\usepackage{booktabs}
\usepackage[dvipsnames]{xcolor}
\usepackage{amssymb}
\usepackage{pifont}
\usepackage{listings}
\usepackage{times}

\usepackage{etoolbox}
\usepackage{multirow}
\usepackage{tabularx}
\usepackage{sectsty}
\usepackage{wrapfig}
\AtBeginEnvironment{quote}{\par\singlespacing\small}

\newcommand{\cmark}{{\color{ForestGreen}\ding{51}}}%
\newcommand{\xmark}{{\color{Maroon}\ding{55}}}%

\definecolor{codegreen}{rgb}{0,0.6,0}
\definecolor{codegray}{rgb}{0.5,0.5,0.5}
\definecolor{codepurple}{rgb}{0.58,0,0.82}
\definecolor{backcolour}{rgb}{0.95,0.95,0.92}
\definecolor{superlightgray}{RGB}{242,242,242}

\lstdefinestyle{mystyle}{
    backgroundcolor=\color{backcolour},
    commentstyle=\color{codegreen},
    keywordstyle=\color{magenta},
    numberstyle=\tiny\color{codegray},
    stringstyle=\color{codepurple},
    basicstyle=\ttfamily\footnotesize,
    breakatwhitespace=false,
    breaklines=true,
    captionpos=b,
    keepspaces=true,
    numbers=left,
    numbersep=5pt,
    showspaces=false,
    showstringspaces=false,
    showtabs=false,
    tabsize=2
}
\newcommand{\numAugraphyAugmentations}{\emph{26}}

\begin{document}

\title{Augraphy: A Data Augmentation Library for\\ Document Images}
\author{Alexander Groleau\inst{1} \and
Kok Wei Chee\inst{1} \and
Stefan Larson\inst{2} \and
Samay Maini\inst{1} \and
Jonathan Boarman\inst{1}}

\authorrunning{A. Groleau et al.}

\institute{Sparkfish LLC (\email{augraphy@sparkfish.com})
\and
Vanderbilt University
}


\maketitle

\begin{abstract}
This paper introduces \emph{Augraphy},\footnote{\url{https://github.com/sparkfish/augraphy}} a Python library for constructing data augmentation pipelines which produce distortions commonly seen in real-world document image datasets.
\emph{Augraphy} stands apart from other data augmentation tools by providing many different strategies to produce augmented versions of clean document images that appear as if they have been altered by standard office operations, such as printing, scanning, and faxing through old or dirty machines, degradation of ink over time, and handwritten markings.
This paper discusses the \emph{Augraphy} tool, and shows how it can be used both as a data augmentation tool for producing diverse training data for tasks such as document denoising, and also for generating challenging test data to evaluate model robustness on document image modeling tasks.

\end{abstract}

\section{Introduction and Motivation}
Daily life in the modern world is increasingly data-driven, involving myriad tasks which require the generation and handling of unstructured data.
Increasingly, this data occurs in the form of messages or images, often with complex contents.
In the office space, most data is created and stored document-form, easy to print and distribute.
The lifetime of these documents typically includes physical alterations by printing, scanning, or photocopying operations, so real documents may appear visually very different from their original born-digital counterparts.
Real-world processes can introduce many types of distortions: folds, wrinkles, or tears in a page can cause color changes and shadows in a scanned document image; low or high printer ink settings may cause some regions of a document to be lighter or darker; human-made annotations like highlighting or pencil marks can add noise to the page.

The outcomes of many machine learning operations involving documents are impacted by the presence of such noise; high-level tasks like document classification and information extraction are frequently expected to perform well even on noisily-scanned document images.
For instance, the RVL-CDIP document classification corpus \cite{harley2015icdar-rvlcdip} consists of scanned document images, many of which have substantial amounts of scanner-induced noise, as does the FUNSD form understanding benchmark \cite{jaume2019-funsd}.
Other intermediate-level tasks like optical character recognition (OCR) and page layout analysis may perform optimally if noise in a document image is minimized \cite{character-recognition-systems,ogorman-document-image-analysis,Rotman2022-hh}.
The lower-level task of document denoising tackles the document noise problem directly, by attempting to remove noise from a document image \cite{ref_noisyoffice,blind-denoising-iccv-2021,kulkarni-2020,patch-based-document-denoising,Mustafa_2018-wan}.
All of these tasks benefit from copious amounts of training data, and one way of generating large amounts of training data with noise-like artefacts is to use data augmentation, reducing the need for human toil.

\begin{figure}
    \centering

    \begin{tabularx}{\linewidth}{*{3}{>{\centering\arraybackslash}X}}
        Noisy Original & Clean Reproduction & \emph{Augraphy} Reproduction \\
        \multicolumn{3}{c}{
            \resizebox{\textwidth}{!}{
                \includegraphics{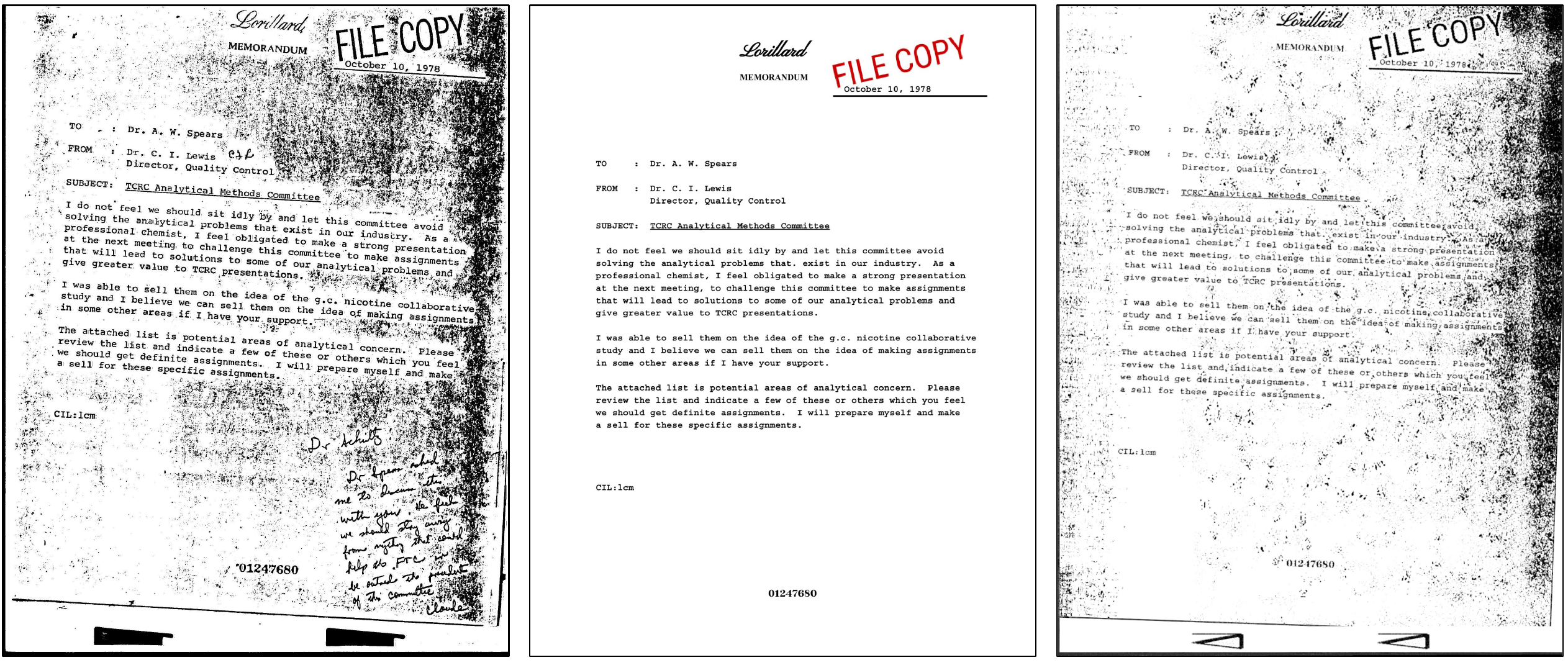}}}      \\
    \end{tabularx}

    \caption{\emph{Augraphy} can be used to introduce noisy perturbations to document images like the original noise seen above, which is a real-life sample from RVL-CDIP. We re-created a clean reproduction and then created a noisy version by applying several \emph{Augraphy} augmentations.}

    \label{fig:intro}
\end{figure}

For this reason we introduce \emph{Augraphy}, an open-source Python-based data augmentation library for generating versions of document images that contain realistic noise artefacts commonly introduced via scanning, photocopying, and other office procedures.
\emph{Augraphy} differs from most image data augmentation tools by specifically targeting the types of alterations and degradations seen in document images.
\emph{Augraphy} offers \numAugraphyAugmentations ~individual augmentation methods out-of-the-box across three ``phases" of augmentations, and these individual phase augmentations can be composed together along with a ``paper factory" step where different paper backgrounds can be added to the augmented image.
The resulting images are realistic, noisy versions of clean documents, as evidenced in Figure~\ref{fig:intro}, where we apply several \emph{Augraphy} augmentations to a clean document image in order to mimic the types of noise seen in a real-world noisy document image from RVL-CDIP.

\emph{Augraphy} has been used in several research efforts: Larson et al. (2022) \cite{larson-2022-rvlcdip-ood} used \emph{Augraphy} to mimic scanner-like noise for evaluating document classifiers trained on RVL-CDIP; Jadhav et al 2022 \cite{jadhav2022pix2pix} used \emph{Augraphy} to generate noisy document images for training a document denoising GAN; and Kim et al. 2022 \cite{webvicob-2022-naver} used \emph{Augraphy} as part of a document generation pipeline for document understanding tasks.

This paper provides an overview of the \emph{Augraphy} library, and demonstrates how it can be used both as a training data augmentation tool and as an effective means for producing data for robustness testing.

\begin{table}[]
    \centering
    \caption{Comparison of various image-based data augmentation libraries. Number of augmentations is a rough count, and many augmentations in other tools are what \emph{Augraphy} calls Utilities. Further, many single augmentations in \emph{Augraphy} --- geometric transforms, for example --- are represented by multiple classes in other libraries.}
    \resizebox{0.9\textwidth}{!}{
    \begin{tabular}{lccccc}
    \toprule
         & \textbf{Number of} & \textbf{Document} & \textbf{Pipeline} & & \\
        \textbf{Library} & \textbf{Augmentations} & \textbf{Centric} & \textbf{Based} & \textbf{Python} & \textbf{License}  \\
    \midrule
        \emph{Augmentor} \cite{augmentor} & 27 & \xmark & \cmark  & \cmark & MIT \\
        \emph{Albumentations} \cite{ref_albumentations} & 216 & \xmark & \cmark & \cmark & MIT \\
        \emph{imgaug} \cite{ref_imgaug} & 168 & \xmark & \cmark & \cmark & MIT \\
        \emph{Augly} \cite{Papakipos2022-gq-augly} & 34 & \xmark & \cmark & \cmark & MIT \\
        \emph{DocCreator} \cite{ref_DocCreator} & 12 & \cmark & \xmark & \xmark & LGPL-3.0 \\
        \emph{Augraphy} (ours) & \numAugraphyAugmentations & \cmark & \cmark & \cmark & MIT\\
    \bottomrule
    \end{tabular}}
    \label{tab:augmentation-library-comparison}
\end{table}

\section{Related Work}

This section discusses prior work related to data augmentation and robustness testing, especially as it relates to document understanding and processing tasks.

\subsection{Data Augmentation}
A wide variety of data augmentation tools and pipelines exist for machine learning tasks, including  natural language processing (e.g., \cite{feng-etal-2021-survey,fadaee-etal-2017-data,wei-zou-2019-eda}), audio and speech processing (e.g., \cite{ko15_interspeech,audiogmenter,audio-framework}), and computer vision and image processing.
In the latter realm, popular tools include \emph{Augly} \cite{Papakipos2022-gq-augly}, \emph{Augmentor} \cite{augmentor}, \emph{Albumentations} \cite{ref_albumentations}, \emph{DocCreator} \cite{ref_DocCreator}, and \emph{imgaug} \cite{ref_imgaug}.
Augmentation strategies from these image-centric libraries are typically general-purpose, and include transformations like rotations, warps, and color modifications.
Table~\ref{tab:augmentation-library-comparison} compares \emph{Augraphy} with other image augmentation libraries and tools.
As can be seen, these other data augmentation libraries do not specifically provide support for imitating the corruptions commonly seen in document analysis corpora.

A notable exception to this is the \emph{DocCreator} image synthesizing tool \cite{ref_DocCreator}, which is targeted towards creating synthetic images that mimic common corruptions seen in document collections.
\emph{DocCreator} differs from \emph{Augraphy} in several crucial ways.
The first difference is that \emph{DocCreator}'s augmentations are meant to imitate those seen in historical (e.g., ancient or medieval) documents, while \emph{Augraphy} is intended to replicate noise caused by office room procedures.
\emph{DocCreator} was also written in the C++ programming language as a monolithic what-you-see-is-what-you-get tool, and does not have a scripting or API interface to enable use in a broader machine learning pipeline.
\emph{Augraphy}, in contrast, is written in Python and can be easily integrated into machine learning model development and evaluation pipelines, alongside other Python packages like PyTorch \cite{ref_pytorch}.

\subsection{Robustness Testing}

The introduction of noise-like corruptions and other modifications to image data can be used as a way of estimating and evaluating model robustness.
Prior work in this space includes the use of image blurring, contrast and brightness changes, color alterations, partial occlusions, geometric transformations, pixel-level noise (e.g., salt-and-pepper noise, impulse noise, etc.), and compression artefacts (e.g., JPEG) to evaluate image classification and object detection models (e.g., \cite{image-quality-impact,imagenet-c,pathology-recommendations,Hosseini2017-kn-google-api,face-recognition-impact,pathology-schomig,Vasiljevic2016-al-bluring-impact}).
More specific to the document understanding field, recent prior work has used basic noise-like corruptions to evaluate the robustness of document classifiers trained on RVL-CDIP \cite{saifullah-2022}.
Our paper also uses robustness testing as a way to showcase the effectiveness of \emph{Augraphy}, but rather than general image modifications like those described above, we use document-centric modifications to produce human-readable distortions which challenge OCR models.

\section{Document Distortion, Theory \& Technique}
Many approaches exist for adding features to an image, and many types of features can be generated. The most common types of features added are Gaussian noise, blurring, geometric transformations like scaling, rotating, translating, and cropping, downsampling, font weighting, and so on.

These types of feature are certainly useful in general image analysis and understanding, but bear little relation to the types of features commonly found in real-world documents.

\emph{Augraphy}'s suite of augmentations was designed to faithfully reproduce the level of complexity in the document lifecycle at seen in the real world.  Such real-world conditions and  features have direct implementations either within the library or on the development roadmap.

Some techniques exist for introducing these features into images of documents, including but not limited to the following:
\begin{enumerate}
\item Text can be generated independently of the paper texture, and can be overlaid onto the ``paper'' by a number of blending functions, allowing a variety of paper textures to be used.
\item Similarly, any markup features may be generated and overlaid by the same methods.
\item Documents can be digitized with a commercial scanner, or converted to a continuous analog signal and back with a fax machine.
\item The finished document image can be used as a texture and attached to a 3D mesh, then projected back to 2 dimensions to simulate physical deformation. DocCreator's 3D mesh support here is excellent and has inspired \emph{Augraphy} to add similar functionality in its roadmap.  In the meantime, \emph{Augraphy} currently offers support for paper folds.
\end{enumerate}

\section{Augraphy}
This section dives into detail about the \emph{Augraphy} library.
We first provide a high-level overview of the library, then discuss the various augmentations supported out-of-the-box, and finally discuss details of the structure of the library.

\subsection{Overview}

\emph{Augraphy} is a lightweight Python package for applying realistic perturbations to document images. It is registered on the Python Package Index (PyPI) and can be installed simply using \colorbox{superlightgray}{\textbf{\texttt{pip install augraphy}}}.

\emph{Augraphy} requires only a few commonly-used Python scientific computing or image handling packages, such as NumPy \cite{ref_numpy}, and has been tested on Windows, Linux, and Mac computing environments and supports recent major versions of Python 3.  Below is a basic out-of-the-box \emph{Augraphy} pipeline demonstrating its usage:

\begin{lstlisting}[language=Python, label={lst:sample-code}, caption={Augmenting a document image with \emph{Augraphy}.}]
import augraphy; import cv2
pipeline = augraphy.default_augraphy_pipeline()
img = cv2.imread("image.png")
data = pipeline.augment(img)
augmented_img = data["output"]
\end{lstlisting}

Modern frameworks for machine learning like \emph{fastai} \cite{ref_fastai} aim to simplify the data handling requirements, and concordantly, the \emph{Augraphy} development team takes great pains to ensure our library's ease-of-use and compatibility.

\emph{Augraphy} is designed to be immediately useful with little effort, especially as part of a preprocessing step for training machine learning models, so great care was taken to establish good defaults.
The default \emph{Augraphy} pipeline (shown in the code snippet above) makes use of all of the augmentations available in \emph{Augraphy}, with starting parameters selected after manual visual inspection of several thousand images.

\emph{Augraphy} provides \numAugraphyAugmentations ~unique augmentations, which may be sequenced into pipeline objects which carry out the image manipulation.
Users of the library can define directed acyclic graphs of images and their transformations via the AugraphyPipeline API, representing the passage of a document through real-world alterations.

\begin{figure}
\centering
\scalebox{0.3}{
\includegraphics[]{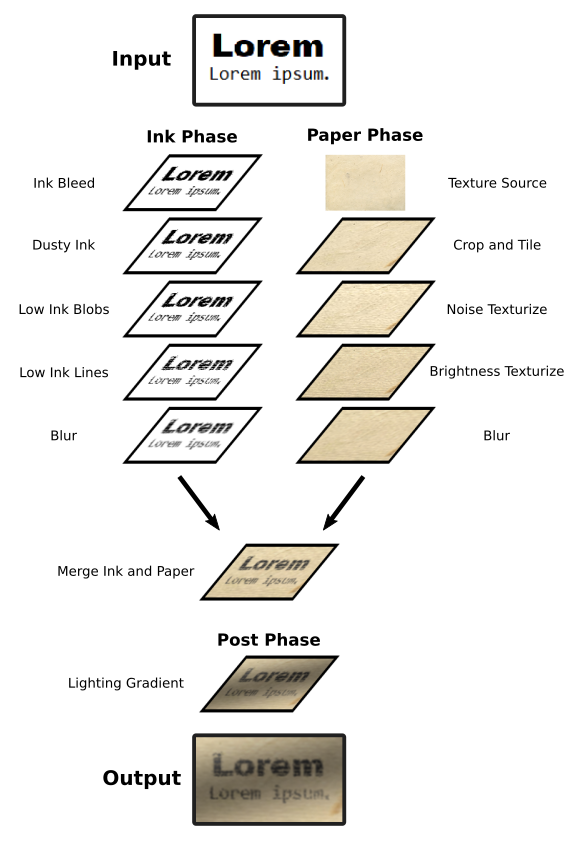}}
\caption{Visualization of an \emph{Augraphy} pipeline, showing the composition of several image augmentations together with a specific paper background}
\label{fig:pipeline}
\end{figure}

\emph{Augraphy} attempts to decompose the lifetime of features accumulating in a document by separating the pipeline into three phases: ink, paper, and post.
The ink phase exists to sequence effects which specifically alter the printed ink --- like bleedthrough from too much ink on page, extraneous lines or regions from a mechanically faulty printer, or fading over time --- and transform them prior to ``printing".

The paper phase applies transformations of the underlying paper on which the ink gets printed; here, a \texttt{PaperFactory} generator creates a random texture from a set of given texture images, as well as effects like random noise, shadowing, watermarking, and staining.
After the ink and paper textures are separately computed, they are merged in the manner of Technique 1 from the previous section, simulating the printing of the document.

After ``printing", the document enters the post phase, wherein it may undergo modifications that might alter an already-printed document out in the world.
Augmentations are available here which simulate the printed page being folded along multiple axes, marked by handwriting or color highlighter, faxed, photocopied, scanned, photographed, burned, stained, and so on.
Figure~\ref{fig:pipeline} shows the individual phases of an example pipeline combining to produce a noised document image.

\begin{figure}
    \centering\scalebox{0.32}{
    \includegraphics{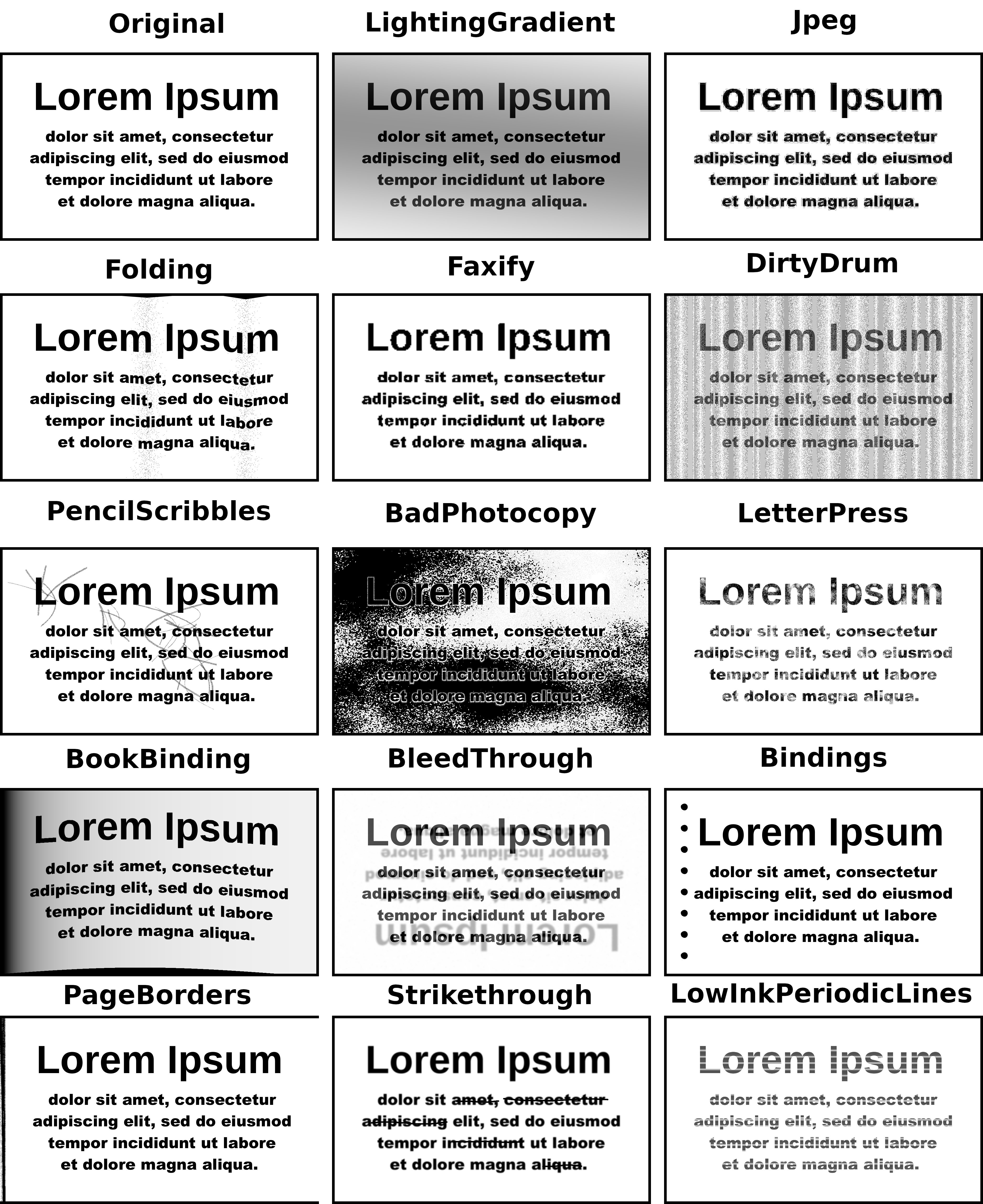}}
    \caption{Examples of some \emph{Augraphy} augmentations.}
    \label{fig2}
\end{figure}

\subsection{Augraphy Augmentations}

\emph{Augraphy} provides \numAugraphyAugmentations ~out-of-the-box augmentations, listed in Table~\ref{tab:augmentations}, with select examples shown in Figure~\ref{fig2}.
As mentioned before, Ink Phase augmentations include those that imitate noisy processes that occur in a document's life cycle when ink is printed on paper.
These augmentations include \texttt{BleedThrough}, which imitates what happens when ink bleeds through from the opposite side of the page.
Another, \texttt{LowInkLines}, produces a streaking behavior common to printers running out of ink.

Augmentations provided by the Paper Phase include \texttt{BrightnessTexturize}, which introduces random noise in the brightness channel to emulate paper textures, and \texttt{Watermark}, which imitates watermarks in a piece of paper.
Finally, the Post Phase includes augmentations that imitate noisy-processes that occur after a document has been created.
Here we find \texttt{BadPhotoCopy}, which uses added noise to generate an effect of dirty copier, and \texttt{BookBinding}, which creates an effect with shadow and curved lines to imitate how a page from a book might appear after capture by a flatbed scanner.

\begin{table}[]
    \centering
    \caption{Individual \emph{Augraphy} augmentations for each augmentation phase, in suggested position within the pipeline. Augmentations that work well in more than one phase are listed in the last column.}
    \scalebox{0.9}{
    \begin{tabular}{llll}
    \toprule
        \textbf{Ink Phase} & \textbf{Paper Phase} & \textbf{Post Phase} & \textbf{Multiple} \\
        \midrule
        \textsf{BleedThrough} &  \textsf{ColorPaper}& \textsf{BadPhotoCopy} & \textsf{BrightnessTexturize} \\
        \textsf{LowInkLines} & \textsf{Watermark} & \textsf{BindingsAndFasteners} & \textsf{DirtyDrum} \\
        \textsf{InkBleed} & \textsf{Gamma} & \textsf{BookBinding} & \textsf{DirtyRollers} \\
        \textsf{Letterpress} & \textsf{LightingGradient} & \textsf{Folding} & \textsf{Dithering} \\
        & \textsf{SubtleNoise} & \textsf{JPEG} & \textsf{Geometric}\\
        & & \textsf{Markup} & \textsf{NoiseTexturize}\\
        & & \textsf{Faxify} & \textsf{PencilScribbles}\\
        & & \textsf{PageBorder} & \\
        \bottomrule
    \end{tabular}}
    \label{tab:augmentations}
\end{table}

Other general-purpose augmentation libraries already exist for adding basic effects like blur, scaling, and rotation, but \emph{Augraphy} also includes these types of augmentations for completeness and utility.
Descriptions of all \emph{Augraphy} augmentations are available online, along with the motivation for their development and usage examples. \footnote{\url{https://augraphy.readthedocs.io/en/latest/}}

\subsection{The Library}
\emph{Augraphy} is a Python-based library, allowing for maximal accessibility for practitioners, and is designed with an object-oriented structure, with concerns divided across a class hierarchy.
When composed, augmentations from the library interact to produce complex document image transformations, generating realistic new synthetically-augmented document images.

There are four ``main sequence" classes in the \emph{Augraphy} codebase, which together provide the bulk of the library's functionality:

\smallskip
\noindent\textbf{\texttt{Augmentation}.} ~The \texttt{Augmentation} class is the most basic class in the project, and essentially exists as a thin wrapper over a probability value in the interval [0,1].
Every augmentation contained in a pipeline has its own chance of being applied during that pipeline's execution.

\smallskip
\noindent\textbf{\texttt{AugmentationResult}.} ~After an augmentation is applied, the output of its execution is stored in an \texttt{AugmentationResult} object and passed forward through the pipeline.
These also record a full copy of the augmentation runtime data, as well as any metadata that might be relevant for debugging or other advanced use.

\smallskip
\noindent\textbf{\texttt{AugmentationSequence}.} ~A list of \texttt{Augmentation}s --- together with the intent to apply those Augmentations in sequence --- determines an \texttt{AugmentationSequence}, which is itself both an \texttt{Augmentation} and callable (behaves like a function).
In practice, these are the model for the pipeline phases discussed previously; they are essentially lists of \texttt{Augmentation} constructor calls which produce callable \texttt{Augmentation} objects of the various flavors explored in \texttt{AugmentationSequences} are applied to the image during each of the \texttt{AugmentationPipeline} phases, and in each case yield the image, transformed by some of the Augmentations in the sequence.

\smallskip
\noindent\textbf{\texttt{AugmentationPipeline}.} ~The bulk of the heavy lifting in \emph{Augraphy} resides in the Augmentation pipeline, which is our abstraction over one or more events in a physical document's life.
Consider the following sequence of events:
\begin{enumerate}
  \item Ink is adhered to the paper material during the initial printing of a document.
  \item The document is attached to a public bulletin board in a high-traffic area.
  \item Over a period of several weeks, it is annotated, defaced, and eventually torn away from its securing staples, flying away in the wind.
  \item Fifty years later, our protagonist page is discovered by cleaning staff, who turn it over to library archivists.
  \item These conservationists use delicate tools to gently position and record an image of the document, storing this in a digital repository.
\end{enumerate}
An \texttt{AugmentationPipeline} can model this entire sequence of events, or any individual event within.

Realistically reproducing effects in document images requires careful attention to how those effects are produced in the real world.
Many issues, like the various forms of misprint, only affect text and images on the page.
Others, like a coffee spill, change properties of the paper itself. Further still, there are transformations like physical deformations which alter the geometry and topology of both the page material and the graphical artifacts on it.
Effectively capturing processes like these in reproducible augmentations means separating our model of a document pipeline into ink, paper, and post-processing layers, each containing some augmentations that modify the document image as it passes through.
Producing realistically noisy document images can now be reduced to the definition and application of one or more \emph{Augraphy} pipelines to some clean, born-digital document images.

The value added by the \texttt{AugraphyPipeline} class over a bare list of functions applied in sequence to an image is principally in the collection of runtime metadata; the output of an \texttt{AugraphyPipeline} application is a Python dictionary which contains not only the final image, but copies of every intermediate image, as well as information about the object constructors and their parameters that were used for each augmentation.
 This pipeline metadata allows for easy inspection and fine-tuning of the pipeline definition to achieve outputs with desired features, facilitating (re)production of documents as in Figure \ref{fig:intro}.

There are also two classes that provide additional critical functionality in order to round out the \emph{Augraphy} base library:

\noindent\textbf{\texttt{OneOf}.} ~To model the possibility that a document image has undergone one and only one of a collection of augmentations, we use \texttt{OneOf}, which simply selects one of those augmentations from the given list, and uses this to modify the image.

\noindent\textbf{\texttt{PaperFactory}.} ~We often print on multiple sizes and kinds of paper, and out in the world we certainly \textit{encounter} such diverse documents.
We introduce this variation into the \texttt{AugmentationPipeline} by including \texttt{PaperFactory} in the \texttt{paper} phase of the pipeline.
This augmentation checks a local directory for images of paper to crop, scale, and use as a background for the document image.
The pipeline contains edge detection logic for lifting only text and other foreground objects from a clean image, greatly simplifying the ``printing" onto another ``sheet", and capturing in a reproducible way the construction method used to generate the NoisyOffice database \cite{ref_NoisyOfficeDatabase}.
Taken together, \texttt{PaperFactory} makes it trivial to re-print a document onto other surfaces, like hemp paper, cardboard, or wood.

\smallskip
Interoperability and flexibility are core requirements of any data augmentation library, so \emph{Augraphy} includes several utility classes designed to improve developer experience:

\noindent\textbf{\texttt{ComposePipelines}.} ~This class provides a means of composing two pipelines into one, allowing for the construction of complex multi-pipeline operations.

\smallskip
\noindent\textbf{\texttt{Foreign}.} ~This class can be used to wrap augmentations from external projects like \emph{Albumentations} and \emph{imgaug}.

\smallskip
\noindent\textbf{\texttt{OverlayBuilder}.} ~This class uses various blending algorithms to fuse foreground and background images together, which is useful for ``printing" or applying other artifacts like staples or stamps.

\smallskip
\noindent\textbf{\texttt{NoiseGenerator}.} ~This class uses make blobs algorithm to generate mask of noises in different shape and location.

\section{Deep Learning with \emph{Augraphy}}
\emph{Augraphy} aims to facilitate rapid dataset creation, advancing the state of the art for document image analysis tasks.
This section describes a brief experiment in which \emph{Augraphy} was used to augment a new collection of clean documents, producing a new corpus which fills a feature gap in existing public datasets. This collection was used to train a denoising convolutional neural network capable of high-accuracy predictions, demonstrating \emph{Augraphy}'s utility for robust data augmentation.

All code used in these experiments is available in the \texttt{augraphy-paper} GitHub repository \footnote{\url{https://github.com/sparkfish/augraphy-paper}}.

\subsection{Model Architecture}
To evaluate \emph{Augraphy}, we used an off-the-shelf Nonlinear Activation-Free Network (NAFNet), \cite{ref_nafnet}, making only minor alterations to the model's training hyperparameters, changing the batch size and learning epoch count to fit our training data.

\subsection{Data Generation}
Despite recent techniques (\cite{ref_tvt2040}, \cite{ref_vtssds}) for reducing the volume of input data required to train models, large datasets remain king; feeding a model more data during training can help ensure better latent representations of more features, improving robustness of the model to noise and increasing its capacity for generalization to new data.

To train our model, we developed the \emph{ShabbyPages} dataset \cite{ref_ShabbyPages,shabbypages_paper}, a real-world document image corpus with \emph{Augraphy}-generated synthetic noise.
Gathering the 600 ground-truth documents was the most challenging part, requiring the efforts of several crowdworkers across multiple days. Adding the noise to these images was trivial however; \emph{Augraphy} makes it easy to produce large training sets.
For additional realistic variation, we also gathered 300 paper textures and used the \texttt{PaperFactory} augmentation to ``print'' our source documents onto these.

The initial 600 PDF documents were split into their component pages, totaling 6,202 clean document images exported at 150 pixels per inch.
We iteratively developed an \emph{Augraphy} pipeline which produced satisfactory output with few overly-noised images, then ran each of the clean images through this to produce the base training set.
Patches were taken from each of these images and used for the final training collection.

Another new dataset --- \emph{ShabbyReal} --- was produced as part of the larger \emph{ShabbyPages} corpus \cite{ref_ShabbyPages,shabbypages_paper}; this data was manually produced by applying sequences of physical operations to real paper. \emph{ShabbyReal} is fully out-of-distribution for this experiment, and also has a very high degree of diversity, providing good conditions for evaluating \emph{Augraphy}'s effect.

\subsection{Training Regime}
The NAFNet was given a large sample to learn from: for each image in the ShabbyPages set, we sampled ten patches of 400x400 pixels, using these to train the instance across just 16 epochs.

The model was trained with mean squared error as the loss function, using the Adam optimizer and cosine annealing learning rate scheduler, then evaluated with the mean average error metric.

\subsection{Results}
\begin{figure}
\centering\scalebox{1.0}{
\includegraphics{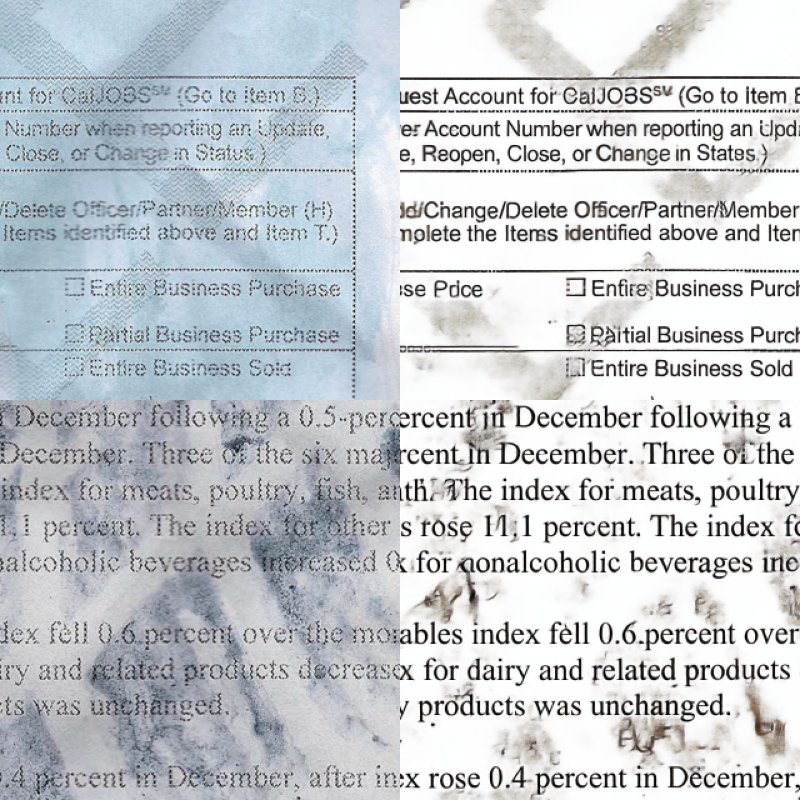}}
\caption{Sample \emph{ShabbyReal} noisy images (left) and their denoised counterparts (right).}
\label{fig:shabbyreal_denoised}
\end{figure}

Sample predictions from the model on the validation task are presented in Figure \ref{fig:shabbyreal_denoised}. The SSIM score indicates that these models generalize very well, though they do overcompensate for shadowing features around text, by increasing the line thickness in the prediction, resulting in a bold font.

To compare the model's performance on the validation task, we considered the root mean square error (RMSE), structural similarity index (SSIM), and peak signal-to-noise ratio (PSNR) metrics.

\begin{table}
\centering
\caption{Denoising performance on \emph{ShabbyReal} validation task}\label{tab1}
\begin{tabular}{@{\hspace{2em}}l@{\qquad}@{\hspace{2em}}l@{\qquad}}
\toprule
\textbf{Metric} & \textbf{Score} \\
\midrule
SSIM & 0.71\\
PSNR & 32.52\\
RMSE & 6.52\\
\hline
\end{tabular}
\end{table}

As can be seen in Figure \ref{fig:shabbyreal_denoised}, the model was able to remove a significant amount of blur, line noise, shadowing from tire tracks, and watercoloring.

\begin{figure}
\centering
\includegraphics[width=0.95\columnwidth]{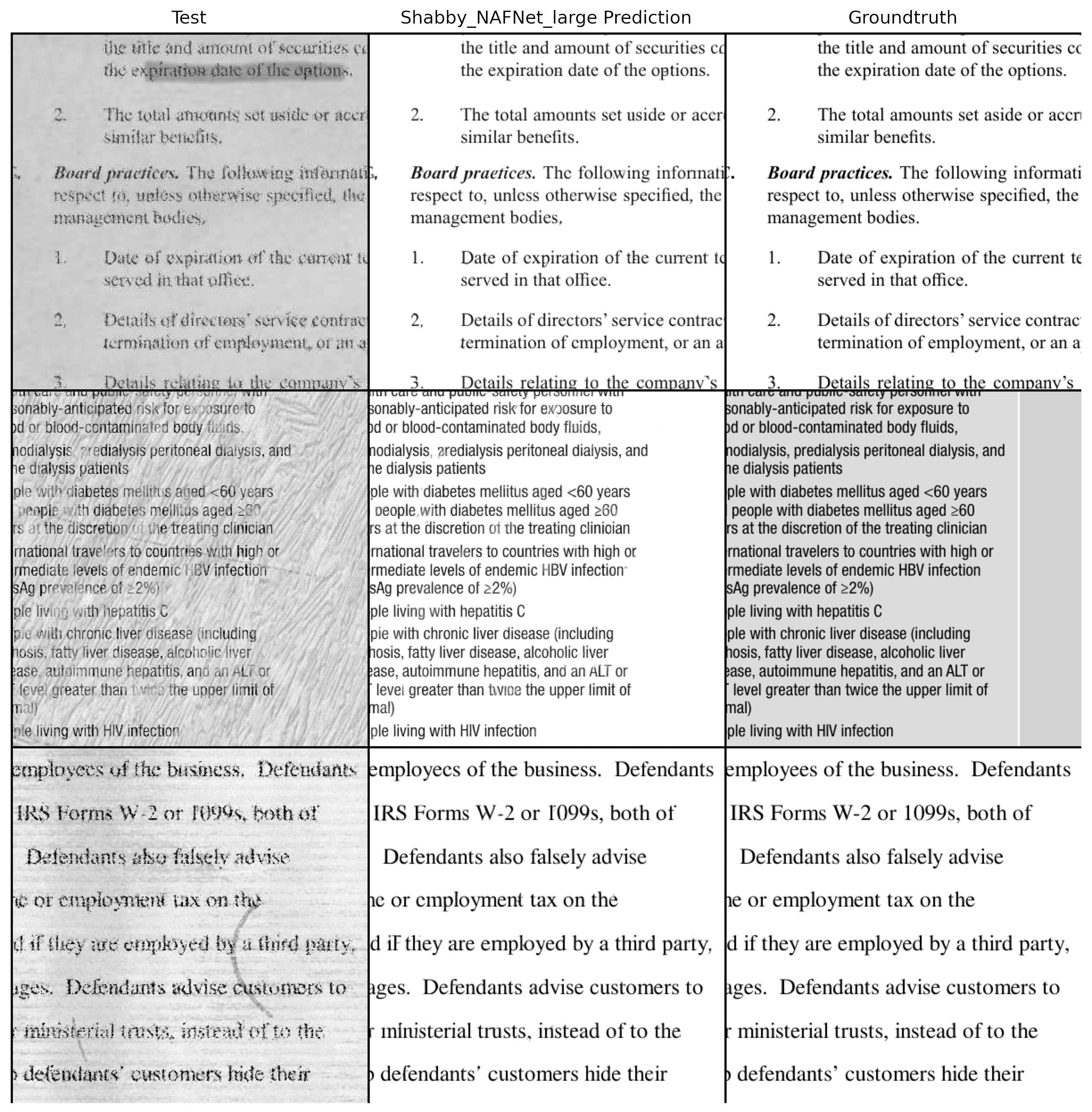}
\caption{Sample \emph{ShabbyPages} noisy images (left), denoised by Shabby\_NAFNet (center), and groundtruth (right).}
\label{fig:shabbypages_denoising}
\end{figure}

\section{Robustness Testing}
In this section, we examine the use of \emph{Augraphy} to add noise which challenges existing models.

We first use \emph{Augraphy} to add noise to an image of text with known groundtruth.
The Tesseract \cite{ref_tesseract} pre-trained OCR model's performance on the clean image is compared to the OCR result on the \emph{Augraphy}-noised image using the Levenshtein distance.

Then, we add \emph{Augraphy}-generated noise to document images which contain pictures of people, and perform face-detection on the output, comparing the detection accuracy to that of the clean images.

\subsection{Character Recognition}
We first compiled 15 ground-truth, noise-free document images from a new corpus of born-digital documents, whose ground-truth strings are known. We then used Tesseract to generate OCR predictions on these noise-free documents, as a baseline for comparison. We considered these OCR predictions as the ground-truth labels for each document. Next, we generated noisy versions of the 15 documents by running them through an \emph{Augraphy} pipeline, and again used Tesseract to generate OCR predictions on these noisy documents. We compared the word accuracy rate on the noisy OCR results versus the ground-truth noise-free OCR results, and found that the noisy OCR results were on average 52\% less accurate, with a range of up to 84\%. This example use-case demonstrates the effectiveness of using \emph{Augraphy} to create challenging test data that remains human readable suitable for evaluating OCR systems.

\begin{figure}
\centering
\includegraphics[width=0.9\columnwidth]{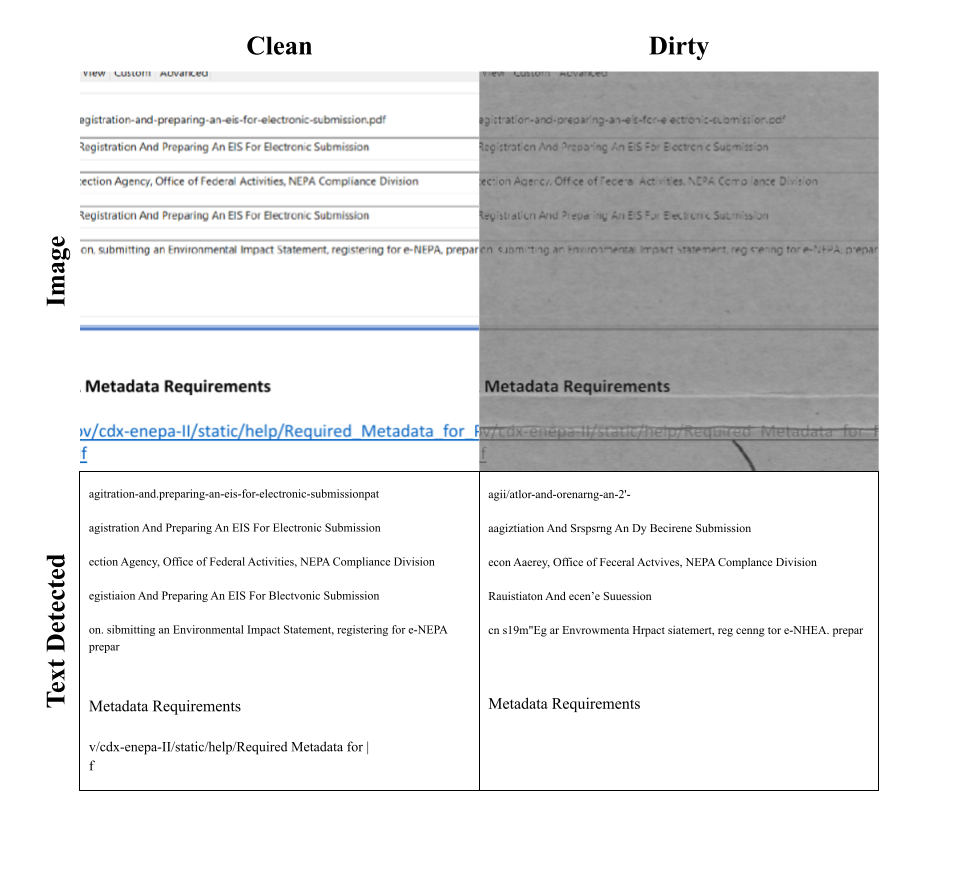}
\caption{Sample \emph{ShabbyReal} clean/noisy image pair, and the OCR output for each. The computed Levenshtein distance between the OCR result strings is 203.}
\label{fig:ocr_output}
\end{figure}

\subsection{Face Detection Robustness Testing}
\begin{wraptable}{r}{6cm}
    \centering
    \caption{Face detection performance.}
    \begin{tabular}{lr}
    \toprule
        \textbf{Model} & \textbf{Accuracy} \\
        \midrule
        Azure & 57.1\%\\
        Google & 60.1\%  \\
        Amazon & 49.1\% \\
        UltraFace & 4.5\%\\
        \bottomrule
    \end{tabular}
    \label{tab:face-detection-results}
\end{wraptable}
In this section, we use \emph{Augraphy} to investigate the robustness of object detection models on images that have been altered by \emph{Augraphy}.
Specifically, we explore using \emph{Augraphy} to add noise to images containing faces in order to test the robustness of face detection models.
In this way, we move beyond analyzing robustness of text-related tasks, but the images we analyze in this section can nonetheless appear in documents like newspapers \cite{newspaper-navigator} and identification documents \cite{midv-500}, which are often scanned from the physical world by noisy scanners.
Hence, it is important for image-focused models like image classifiers and object detection models to be robust to scanner-induced noise.

We begin by sampling 75 images from the FDDB face detection benchmark \cite{fddbTech}.
Then, we use \emph{Augraphy} to generate 10 altered versions of each image and manually remove any augmented image where the face(s) is not reasonably visible.
We then test four face detection models on the noisy and noise-free images.
These models are: proprietary face detection models from Google,\footnote{\url{https://cloud.google.com/vision/docs/detecting-faces}} Amazon,\footnote{\url{https://aws.amazon.com/rekognition/}} and Microsoft\footnote{\url{https://azure.microsoft.com/en-us/services/cognitive-services/face}}, and lastly the freely-available UltraFace\footnote{\url{https://github.com/onnx/models/tree/main/vision/body_analysis/ultraface}} model.

\begin{figure}
    \centering\scalebox{0.47}{
    \includegraphics{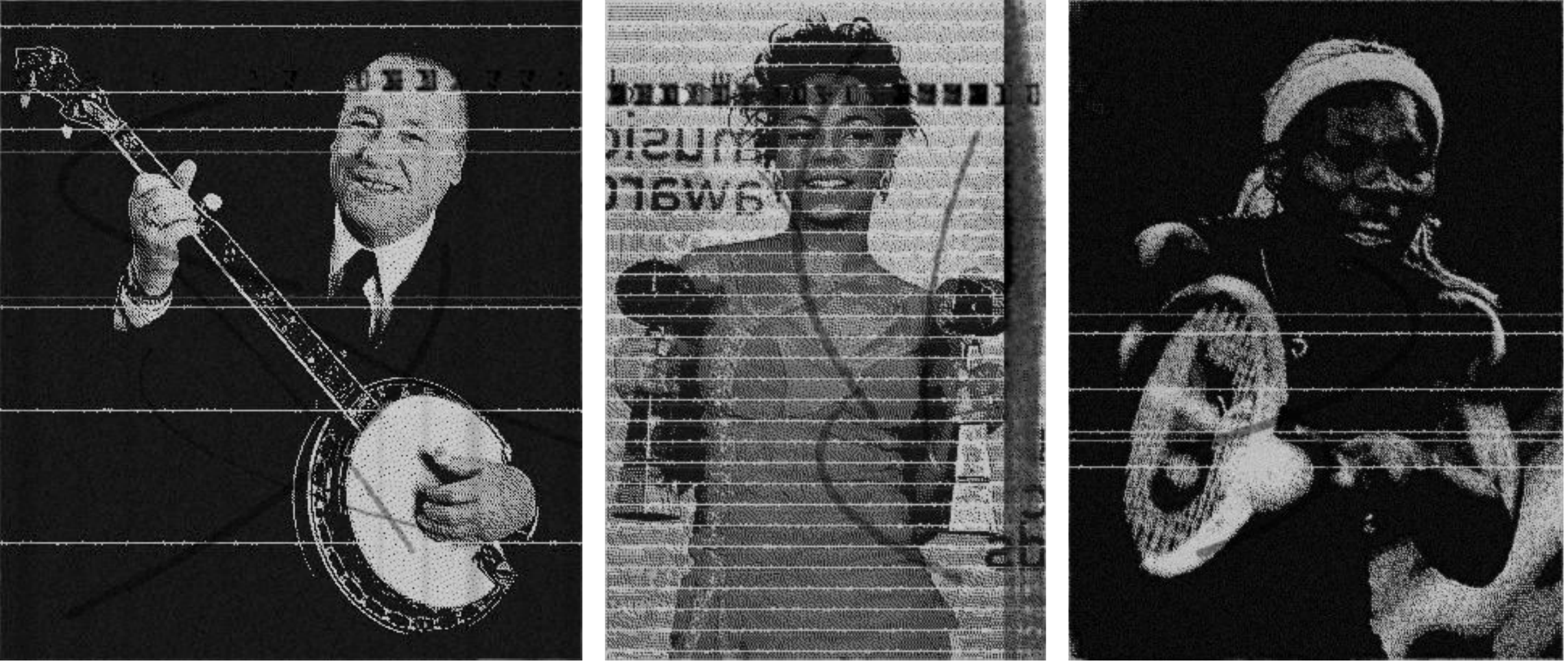}}
    \caption{Example augmented images that yielded false-positive predictions with the Azure face detector.}
    \label{fig:face-detection-mistakes}
\end{figure}

Table~\ref{tab:face-detection-results} shows face detection accuracy of both models on the noisy test set, and example images where no faces were detected by the Microsoft model are shown in Figure \ref{fig:face-detection-mistakes}.
We see that all four models struggle on the \emph{Augraphy}-augmented data, with the proprietary models seeing detection performances drop to between roughly 50-60\%, and UltraFace detecting only 4.5\% of faces that it found in the noise-free data.

\section{Conclusion and Future Work}
We presented \emph{Augraphy}, a framework for generating realistic synthetically-augmented datasets of document images.
Other available image augmentation tools were examined and found to lack features needed for our purposes, motivating the creation of this library specifically targeting the types of alterations and degradations seen in document images.

We described the process for using \emph{Augraphy} to create a new document image dataset containing synthetic real-world noise, then compared results obtained by training a convolutional NAFNet instance on this corpus.

Finally, we examined \emph{Augraphy}'s efficacy in generating confounding data for testing the robustness of document vision models.

Future work on the \emph{Augraphy} library will focus on adding new types of augmentations, increasing performance to enable faster creation of larger datasets on commodity hardware, providing more scale-invariant support so that augmentations perform well at all document image resolutions, and responding to community-initiated feature requests.

\emph{Augraphy} is licensed under the MIT open source license, and readers are invited to share feedback and participate in its development on GitHub.

%
%
%
\bibliographystyle{splncs04}
\bibliography{paper}

\end{document}